\documentclass[conference]{IEEEtran}
\IEEEoverridecommandlockouts
\usepackage{cite}
\usepackage{amsmath,amssymb,amsfonts}
\usepackage[T1]{fontenc}
\usepackage{lmodern}
\usepackage{algorithmic}
\usepackage{graphicx}
\usepackage{textcomp}
\usepackage{mathptmx}
\usepackage{mathtools}
\usepackage{amsmath}
\usepackage{xcolor}

\def\BibTeX{{\rm B\kern-.05em{\sc i\kern-.025em b}\kern-.08em
    T\kern-.1667em\lower.7ex\hbox{E}\kern-.125emX}}
\begin{document}

\title{Differential Privacy-Driven Framework for Enhancing Heart Disease Prediction\\}
\author{
	\IEEEauthorblockN{Yazan Otoum, Amiya Nayak}
	\IEEEauthorblockA{\textit{School of Computer Science and Technology}, 
	\textit{Algoma University}, 
	Canada \\
	Email: \ otoum@algomau.ca
 }
	\IEEEauthorblockA{\textit{School of Electrical Engineering and Computer Science}, 
	\textit{University of Ottawa}, 
	Canada  \\
	Email: \ nayak@uottawa.ca
 }
 }
\maketitle

\footnotetext[1]{© 2025 IEEE. Accepted to IEEE International Conference on Communications ICC 2025. Final version to appear in IEEE Xplore.}

\begin{abstract}

With the rapid digitalization of healthcare systems, there has been a substantial increase in the generation and sharing of private health data. Safeguarding patient information is essential for maintaining consumer trust and ensuring compliance with legal data protection regulations. Machine learning is critical in healthcare, supporting personalized treatment, early disease detection, predictive analytics, image interpretation, drug discovery, efficient operations, and patient monitoring. It enhances decision-making, accelerates research, reduces errors, and improves patient outcomes. In this paper, we utilize machine learning methodologies, including differential privacy and federated learning, to develop privacy-preserving models that enable healthcare stakeholders to extract insights without compromising individual privacy. Differential privacy introduces noise to data to guarantee statistical privacy, while federated learning enables collaborative model training across decentralized datasets. We explore applying these technologies to Heart Disease Data, demonstrating how they preserve privacy while delivering valuable insights and comprehensive analysis. Our results show that using a federated learning model with differential privacy achieved a test accuracy of 85\%, ensuring patient data remained secure and private throughout the process.

\end{abstract}

\section{Introduction}

With the increasing volume of sensitive data, particularly in healthcare, privacy-preserving techniques have become critical for protecting individuals' personal information. One such method, differential privacy, is designed to safeguard privacy by introducing controlled noise to the results of data analyses. This, in turn, preserves the statistical properties of the dataset while concealing individual-level information. This ensures that, even with auxiliary knowledge, an observer cannot confidently infer whether a specific individual’s data is part of the analysis. Differential privacy is governed by a key parameter known as the ``privacy budget'' or ``privacy parameter,'' denoted by \begin{math} \epsilon \end{math} value, which indicates stronger privacy guarantees but may reduce the accuracy of the analysis due to the added noise. There is a well-known trade-off between privacy and data utility—enhancing privacy often decreases the usefulness of the results. An additional parameter, \begin{math} \delta \end{math}, represents the probability of an extreme privacy breach, providing an extra safeguard when the noise is insufficient to protect the data fully. Another emerging solution for privacy-preserving data analysis is federated learning, a decentralized machine learning approach that allows model training on distributed devices—such as smartphones, IoT devices, or edge servers—without transferring raw data to a central server \cite{otoum2024enhancing}. In federated learning, model updates rather than raw data are transmitted to a central server, aggregating them into a global model. This process ensures that sensitive data remains on local devices, reducing the risk of privacy breaches. A prominent example of federated learning in action is predictive text on smartphones. Companies like Apple and Samsung utilize this technology to enhance user experiences without compromising user data privacy. Google’s 2018 paper on federated learning demonstrated how the technology was used to protect the privacy of keyboard users while improving predictive text accuracy \cite{hard2018federated}. The core idea of federated learning is to harness the collective intelligence of a distributed network of devices while maintaining data privacy and ownership. This technique enables collaborative model training across multiple decentralized datasets without centralized data storage, making it a powerful tool for privacy-conscious machine learning applications \cite{otoum2021federated}, \cite{10901705}. In this paper, we explore the integration of differential privacy with federated learning to address privacy concerns in healthcare data analysis. Specifically, we apply these methodologies to heart disease datasets, demonstrating how privacy-preserving techniques can be employed without sacrificing the accuracy of predictive models. Heart disease research requires collaboration across institutions, but the sensitivity of healthcare data demands strong privacy protections. Differential privacy addresses this by enabling the sharing of valuable insights while safeguarding patient information. Combining deep learning with differential privacy helps predict heart disease risks while tackling patient consent, data ownership, and data breaches. Notably, cyberattacks like the LifeLabs breach in 2019 underscore the vulnerability of healthcare data \cite{webster2020canadian}. Our approach integrates federated learning with differential privacy, ensuring data decentralization and minimizing re-identification risks. It also optimizes privacy (epsilon) to maintain model accuracy without compromising privacy, addressing a key limitation of existing methods \cite{said2023scalable}.

The remainder of this paper is structured as follows; Section~\ref{sec:literature_review} reviews related work on federated learning and differential privacy in healthcare. Section~\ref{sec:background} provides an overview of the background and key concepts, including federated learning strategies and privacy mechanisms. Section~\ref{sec:model_dataset} presents the proposed model and datasets, detailing the integration of differential privacy into deep learning for heart disease prediction. Section~\ref{sec:results} discusses experimental results, analyzing the trade-offs between privacy and accuracy. Finally, Section~\ref{sec:con} concludes the paper and outlines future research directions.

\section{Literature Review}
\label{sec:literature_review}

Combining federated learning and differential privacy holds promise in revolutionizing healthcare data analysis. In \cite{xu2021federated}, the authors explore the integration of differential privacy and federated learning to enhance privacy in healthcare data analysis, focusing on heart disease prediction. Federated learning allows decentralized model training while preserving data privacy, and differential privacy ensures that individual patient data remains secure by adding noise to the dataset. Prior studies have applied these methods to electronic health records and predictive modelling. This research builds on those efforts, addressing the growing concern of data breaches and demonstrating how these techniques can ensure data privacy without compromising model accuracy.

The authors in \cite{liu2020secure} address the challenge of preserving privacy in healthcare data by integrating differential privacy with federated learning. Previous studies have highlighted the effectiveness of federated learning for decentralized model training while maintaining data privacy. The authors build on these findings by incorporating differential privacy, which adds noise to ensure individual-level data protection. Prior research has demonstrated these methods’ utility in healthcare, particularly for sensitive data such as heart disease records. The proposed approach aims to enhance privacy without sacrificing model accuracy, contributing to the growing body of work on secure machine learning in healthcare. Additionally, in \cite{li2019privacy}, the authors address the challenge of medical data privacy in federated learning, particularly when training machine learning models on sensitive patient data. Prior research has demonstrated the effectiveness of federated learning in enabling decentralized model training while maintaining privacy. However, the risk of model updates leaking local training data remains. The authors build on previous efforts by integrating differential privacy techniques to enhance data protection. Their evaluation of brain tumour segmentation using the BraTS dataset demonstrates the trade-off between model accuracy and privacy, contributing to the growing body of research on privacy-preserving machine learning.

Lastly, Tang et al. \cite{tang2021personalized} highlight the use of federated learning and differential privacy in electrocardiogram classification, showcasing their effectiveness in delivering accurate diagnoses while protecting patient privacy. This study and others demonstrate how these techniques balance the dual demands of accuracy and privacy in healthcare data analysis, contributing to a significant transformation in medical research. In our work, we aimed to combine differential privacy with federated learning. To implement federated learning, we explored several libraries, including PySyft, PipelineDP, TensorFlow Federated, and Flower. After evaluating these options, we selected Flower due to its broad platform compatibility and support for critical libraries such as TensorFlow and scikit-learn, making it the most suitable choice.

\begin{figure}[ht]
\begin{center}
\includegraphics[keepaspectratio=true,scale=0.5]{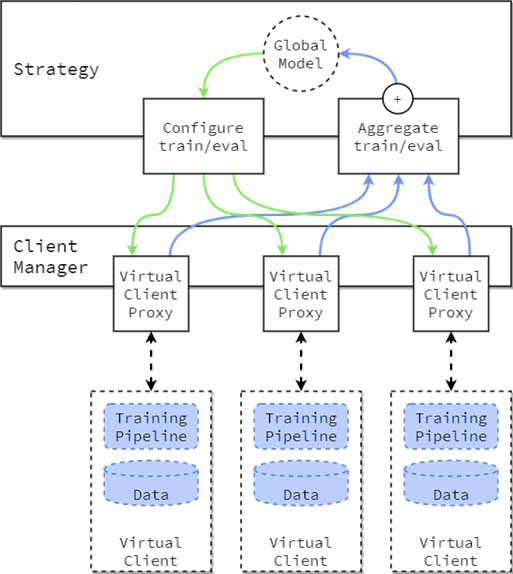}
\caption{Flower Library’s Architecture for Federated Learning \cite{li2021secure}}
\label{fig::Figure1}
\end{center}
\end{figure}

\section{Background}
\label{sec:background}

\begin{figure*}[ht]
\begin{center}
\includegraphics[keepaspectratio=true,scale=1.0]{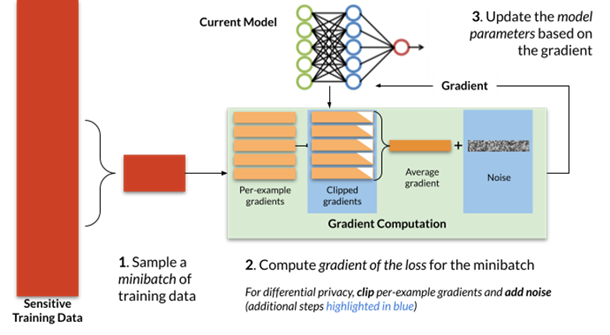}
\caption{Opacus DP Implementation}
\label{fig::Figure2}
\end{center}\end{figure*}

This work used the Flower framework to coordinate and aggregate model updates in a federated learning environment. The Flower framework offers flexibility in implementing various strategies for synchronizing and updating client models, with our implementation leveraging the Federated Averaging (FedAvg) strategy. To enhance privacy, we integrated differential privacy using IBM’s $diffprivlib$ library, which adds noise to model updates, ensuring individual data cannot be identified. Combined with differential privacy, Federated learning protects sensitive data by decentralizing model training and safeguarding individual privacy. In our study, we applied differential privacy with an epsilon value of 1.0, reflecting a balance between privacy protection and model accuracy. Using logistic regression on a heart disease dataset, we achieved an accuracy of 47\%, which we sought to improve by experimenting with different data partitioning techniques across multiple clients. The results were consistent across approaches, leading us to explore deep learning models as a next step. Additionally, we investigated integrating blockchain technology with federated learning to secure data sharing further. Blockchain can enhance the security of federated learning systems by ensuring decentralized, immutable records of transactions, while federated learning ensures sensitive data remains private. However, extending differential privacy to deep learning models proved challenging. After evaluating various differential privacy tools, we selected Opacus, a library compatible with PyTorch, over TensorFlow Privacy due to compatibility issues. Our focus shifted to reanalyzing the dataset and refining our models to achieve better accuracy while maintaining robust privacy guarantees through using Opacus in future deep learning implementations. In addition to leveraging Opacus for differential privacy, we aim to explore multi-party computation (MPC) techniques further to enhance data security in federated learning environments. By combining MPC with blockchain and differential privacy, our framework can offer a comprehensive solution for secure, privacy-preserving data sharing and collaboration across institutions. Future work will also optimize model performance while balancing the trade-offs between privacy, security, and accuracy in large-scale healthcare applications.

\section{The Model and Dataset}
\label{sec:model_dataset}

The models are trained using the Cleveland, UCI heart disease datasets, and an integrated dataset. Differential privacy is applied during training to protect individual data. The model architecture consists of fully connected layers with ReLU activations, dropout for regularization, and a sigmoid output layer for binary classification. Data preprocessing includes one-hot encoding for categorical variables and standardization. The Opacus library ensures privacy by adding noise to gradient updates, with privacy parameters epsilon ($\epsilon$) and delta ($\delta$), balancing privacy and accuracy. Epsilon ($\epsilon$) controls the noise added to protect privacy. Smaller values indicate stronger privacy but potentially reduced accuracy. Delta ($\delta$) represents the probability of a privacy breach occurring, with smaller values indicating a lower chance of identifying any individual’s data. The training process optimizes model weights using stochastic gradient descent, and accuracy vs. epsilon plots visualize the trade-off between privacy and performance. The deep learning model, consisting of five fully connected layers, was integrated with Opacus for differential privacy using the \texttt{PrivacyEngine} class. Experimentation with learning rate, batch size, and epochs helped optimize privacy and accuracy. Categorical encoding for features such as gender and chest pain type improved accuracy by 4\%, and grid search hyperparameter tuning was applied for optimization. The model's robustness was validated using k-fold cross-validation. The introduction of noise during training safeguarded privacy and helped reduce overfitting, improving generalization across different datasets. This ensures that privacy protection does not compromise the model’s ability to generalize to new data, a key concern in healthcare applications. Fig. \ref{fig::Figure2} illustrates the Opacus DP implementation. The \texttt{PrivacyEngine} ensures privacy, and the \texttt{make-private-with-epsilon} method wraps PyTorch training with privacy protection. Key parameters include the dataloader, model, optimizer, target-epsilon, target-delta, and epochs. During training, noise is added to the gradients, with the Privacy Engine monitoring the epsilon value in each loop. Testing evaluates accuracy while maintaining privacy with controlled epsilon values.

\section{Result Analysis}
\label{sec:results}

The next step in our study was to evaluate how the same deep learning model performed on different datasets. Initially, the model was developed for the Cleveland Heart Disease Dataset. However, with minor modifications to the functions, we applied the model to the UCI Heart Disease Dataset, where it exhibited better performance. We created a third dataset by integrating the Cleveland and UCI datasets using deep learning techniques to investigate further. With this integrated dataset, the model showed more consistency in the privacy-accuracy tradeoff curves. The following graphs illustrate the results of varying parameters in the training model using the Cleveland Dataset, the UCI Dataset, and the Integrated Dataset.

\begin{figure}[ht]
\begin{center}
\includegraphics[keepaspectratio=true,scale=0.6]{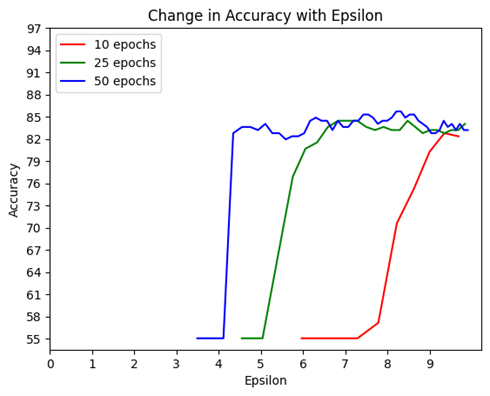}
\caption{Change in accuracy with epsilon in Cleveland Dataset}
\label{fig::Figure3}
\end{center}
\end{figure}

Fig. \ref{fig::Figure3} indicates that using 50 or 25 epochs provides an optimal balance for training the model with differential privacy. At the start of the training process, the privacy engine is informed of the number of epochs, which allows it to manage the trade-off between privacy and accuracy by adjusting the noise levels introduced into the model’s gradients throughout the training. This ensures that the target privacy parameters, such as epsilon, are reached by the end of the specified epochs. In the case of 50 epochs, the model experiences more significant fluctuations in accuracy due to the accumulation of noise over the extended training period. This noise, essential for maintaining privacy, introduces variability in the learning process, resulting in a less stable training curve. However, despite these fluctuations, the model trained with 50 epochs achieves a higher overall accuracy than models trained with fewer epochs. On the other hand, 25 epochs offer a more stable training process with less noise-induced variability. While the final accuracy is slightly lower than that of the 50-epoch model, the smoother training curve indicates that 25 epochs may strike a better balance between stability and performance, particularly when the objective is to reduce overfitting and noise interference.

\begin{figure}[ht]
\begin{center}
\includegraphics[keepaspectratio=true,scale=0.55]{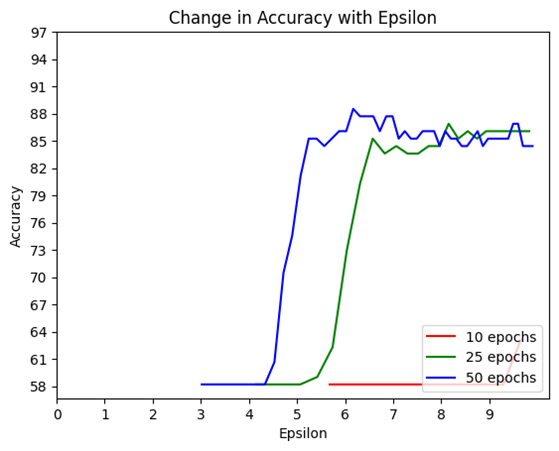}
\caption{Change in accuracy with epsilon in UCI Dataset}
\label{fig::Figure4}
\end{center}\end{figure}

Fig. \ref{fig::Figure4} indicates that 10 epochs are insufficient for effectively training the model on the UCI Heart Disease Dataset. With only 10 epochs, the model does not have enough time to learn meaningful patterns, resulting in suboptimal performance. In contrast, 25 epochs provide a more balanced approach, delivering improved accuracy without significant noise or fluctuations, making it the most stable choice. While 50 epochs offer the highest overall accuracy, the training process is characterized by several dips in accuracy due to the added noise introduced by the differential privacy mechanism. This increased noise can cause the model to overfit or face instability, making 25 epochs a more reliable option.

\begin{figure}[ht]
\begin{center}
\includegraphics[keepaspectratio=true,scale=0.6]{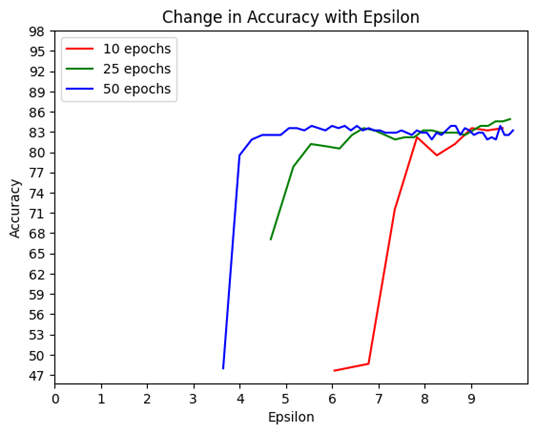}
\caption{Change in accuracy with Epsilon in Integrated Dataset}
\label{fig::Figure5}
\end{center}\end{figure}

We observed a notable improvement in model performance with the integrated dataset, even when trained with only 10 epochs. The additional data provided by integrating the Cleveland and UCI datasets allowed the model to capture patterns better and generalize, enhancing its learning capabilities. As shown in Fig. \ref{fig::Figure5}, the accuracy curve is much smoother, exhibiting fewer dips and fluctuations than models trained on individual datasets. This indicates that integrating datasets improved the training stability and contributed to more consistent accuracy throughout the training process. The number of training loops (epochs) significantly affects both model accuracy and the privacy guarantee, as more epochs allow the model to refine its weights but also introduce more opportunities for noise to accumulate, especially under differential privacy constraints. Our experiments suggest that 25 epochs balance training efficiency and accuracy for this specific dataset. While 50 epochs may yield slightly higher accuracy, it also results in more fluctuations due to noise accumulation, making 25 epochs a more stable and reliable option. Providing the correct number of total epochs to the Privacy Accountant is essential for maintaining this balance. The Privacy Accountant manages the privacy budget (epsilon) throughout the training process. As the number of epochs increases, the privacy budget is gradually consumed, and the noise added to the gradients is adjusted to ensure that privacy guarantees are met. By setting the right number of epochs, we can ensure that the epsilon value is appropriately adjusted, meeting privacy requirements without sacrificing model performance.

\begin{figure}[ht]
\begin{center}
\includegraphics[keepaspectratio=true,scale=0.55]{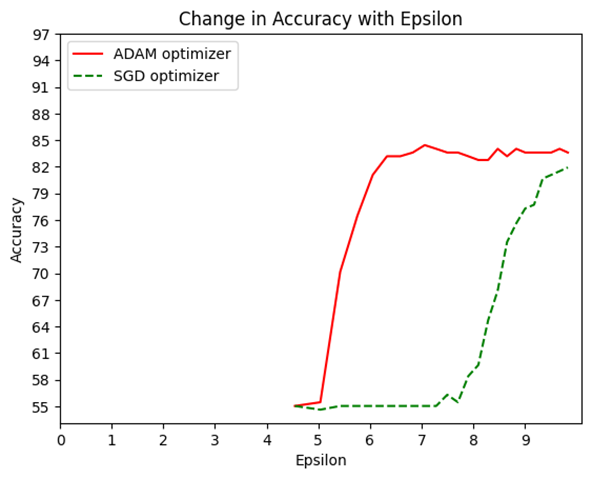}
\caption{Privacy loss with Different optimizers over 25 epochs – Cleveland Dataset}
\label{fig::Figure6}
\end{center}\end{figure}

From Fig. \ref{fig::Figure6}, the model's accuracy gradually improves throughout the training epochs. However, the behaviour of the two optimizers, Adam and SGD, shows a marked difference in how they learn the model. Adam demonstrates a sharper rise in accuracy during the early epochs, indicating that it can quickly adapt to the dataset’s patterns and optimize the model efficiently from the beginning. This rapid convergence results from Adam’s adaptive learning rate mechanism, which adjusts the step size based on the gradients' first and second moments, allowing it to progress faster early in training. In contrast, the SGD optimizer takes longer to begin learning effectively, improving accuracy more slowly in the initial epochs. SGD’s learning process becomes more prominent in the later stages of training as it begins to refine the model's parameters after several epochs. This delayed improvement is due to SGD's fixed learning rate, which makes convergence slower, particularly in the early stages when it requires more iterations to adjust the model’s weights. However, despite its slower start, SGD can still yield strong results given sufficient training time. Overall, while both optimizers ultimately lead to improved accuracy, Adam’s ability to quickly capture the dataset’s characteristics makes it a better choice for faster convergence in the early epochs. At the same time, SGD may perform better in refining the model in the long run, provided there are enough training epochs.

\begin{figure}[ht]
\begin{center}
\includegraphics[keepaspectratio=true,scale=0.55]{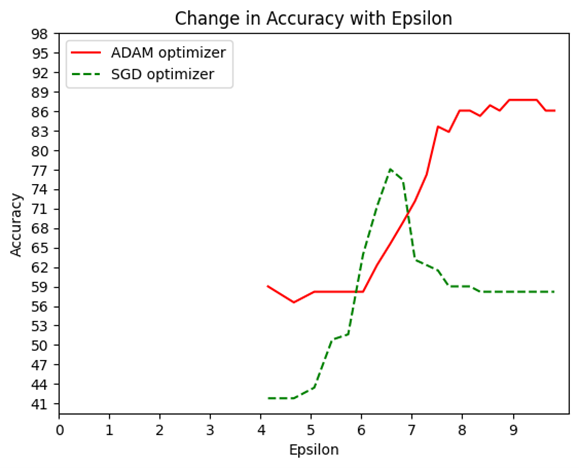}
\caption{Privacy loss with Different optimizers over 25 epochs – Cleveland Dataset}
\label{fig::Figure7}
\end{center}\end{figure}

In Fig. \ref{fig::Figure7}, Adam’s faster convergence is again evident, as it achieves significantly higher accuracy than the SGD optimizer throughout the training process. Adam’s adaptive learning rate, which dynamically adjusts based on the first and second moments of the gradients, allows it to quickly optimize the model parameters, resulting in rapid improvements in accuracy. This ability to adapt and fine-tune the learning rate during training enables Adam to outperform SGD, especially in the early stages. On the other hand, the SGD optimizer, with its fixed learning rate, struggles to reach the same level of accuracy within the same number of epochs. As seen in previous experiments, SGD’s slower convergence requires more epochs to learn the model effectively, making it less efficient in scenarios where quick convergence is crucial. Furthermore, as we observed in earlier graphs, the model trained on the UCI dataset also faced difficulties in achieving high accuracy over 10 epochs. The UCI dataset appears more challenging for the model to learn, as it likely requires more data or better feature representation for effective learning. This may explain why the model performs poorly with both optimizers over short training periods, particularly with SGD. The limited number of epochs exacerbates this issue, as the model does not have sufficient time to fully capture the underlying patterns in the data. Overall, Adam’s ability to adapt quickly to the dataset provides a significant advantage in achieving faster convergence and higher accuracy. At the same time, the slower learning pace of SGD underscores the importance of selecting the right optimizer and training duration based on the dataset’s complexity.

\begin{figure}[ht]
\begin{center}
\includegraphics[keepaspectratio=true,scale=0.55]{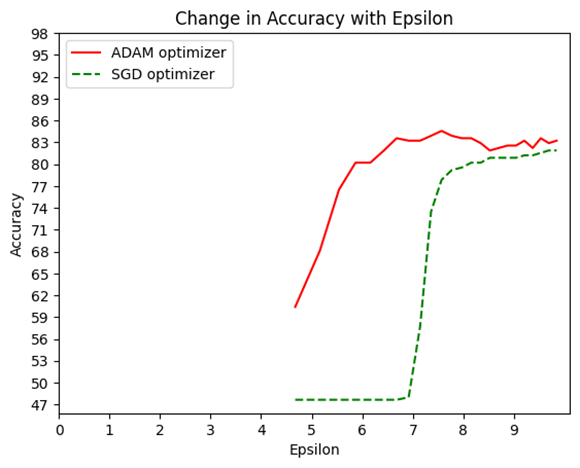}
\caption{Privacy loss with Different optimizers over 25 epochs – Integrated Dataset}
\label{fig::Figure8}
\end{center}\end{figure}

In Fig. \ref{fig::Figure8}, we again observe that the Adam optimizer handles the data most efficiently, consistently outperforming the SGD optimizer. Adam’s ability to adapt its learning rate throughout the training process allows it to converge more quickly and achieve higher accuracy in fewer epochs. This highlights Adam's strength in adjusting to the dataset's characteristics, especially when training deep learning models with complex features. The choice of optimizer plays a critical role in model performance, as evidenced by the differing behaviours across datasets. With its adaptive learning rate mechanism, the Adam optimizer is well-suited for datasets that require rapid convergence and can benefit from dynamic adjustments during training. In contrast, the SGD optimizer, while more stable in the long run, often struggles to capture complex patterns quickly, especially in datasets that demand a more flexible approach. Our results suggest that the optimizer's impact on performance is highly dependent on the nature of the dataset. Different datasets react differently to optimizers, with some benefiting from the faster convergence of Adam, while others may perform better with SGD when given enough epochs to refine the model. This underscores the importance of selecting an appropriate optimizer based on the specific dataset and training objectives, as it can significantly influence the speed and accuracy of the model’s learning process.

\begin{figure}[ht]
\begin{center}
\includegraphics[keepaspectratio=true,scale=0.55]{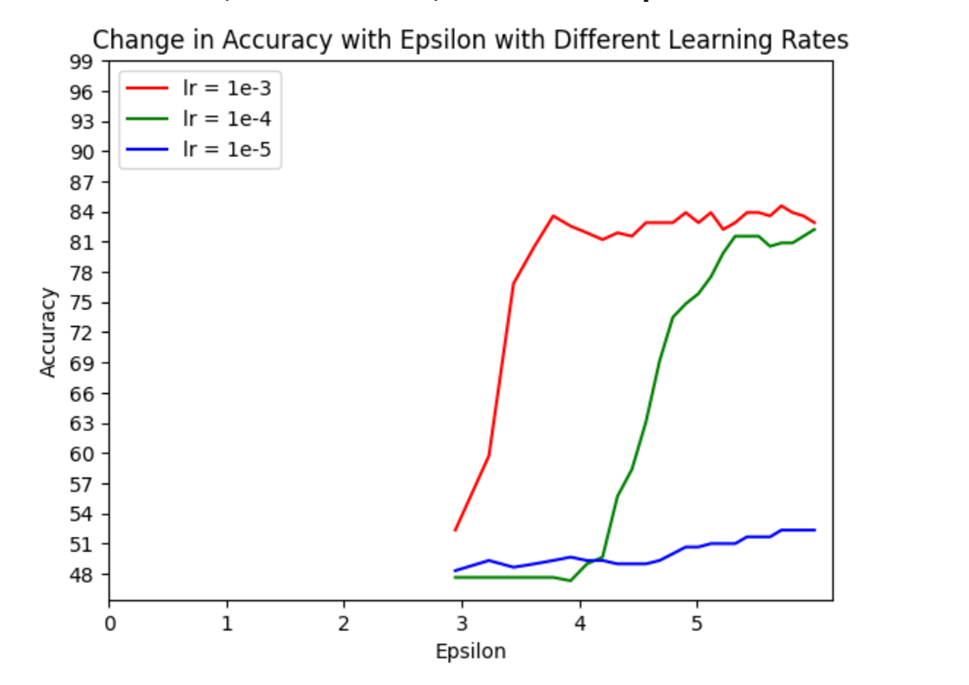}
\caption{Privacy – Accuracy trade-off with different learning rates}
\label{fig::Figure9}
\end{center}\end{figure}

In Fig. \ref{fig::Figure9}, a learning rate of 0.001 is the most suitable, as it strikes the ideal balance between learning efficiency and avoiding convergence to a suboptimal solution. This learning rate allows the model to adjust its parameters quickly enough to capture meaningful patterns without overfitting or excessively fluctuating during training. The epsilon parameter, which signifies the privacy loss factor in differential privacy, is crucial in balancing privacy and model performance. Lower epsilon values guarantee more robust privacy, as they limit the exposure of individual data points. However, as demonstrated in the graphs, aiming for the highest level of privacy (i.e., lower epsilon) often results in a significant decrease in model accuracy. The noise introduced to preserve privacy can obscure important dataset features, leading to less effective learning. An intuitive way to understand the impact of epsilon is through the concept of query sensitivity. Each query made on the dataset reduces the strength of the privacy guarantee, as repeated queries can reveal more information about individual data points, making re-identification more likely. If the number of queries is unlimited, the dataset becomes re-identifiable, compromising privacy. By halving the epsilon parameter—such as reducing it from 10, as used in earlier graphs, to 5 in the subsequent graph—we can effectively double the number of queries that can be answered without compromising privacy. However, this increase in privacy protection comes at the cost of reduced model accuracy, as shown in our results \cite{dyda2021differential}. Striking the right balance between epsilon and learning rate is therefore essential for achieving both strong privacy guarantees and robust model performance.

\begin{figure}[ht]
\begin{center}
\includegraphics[keepaspectratio=true,scale=0.55]{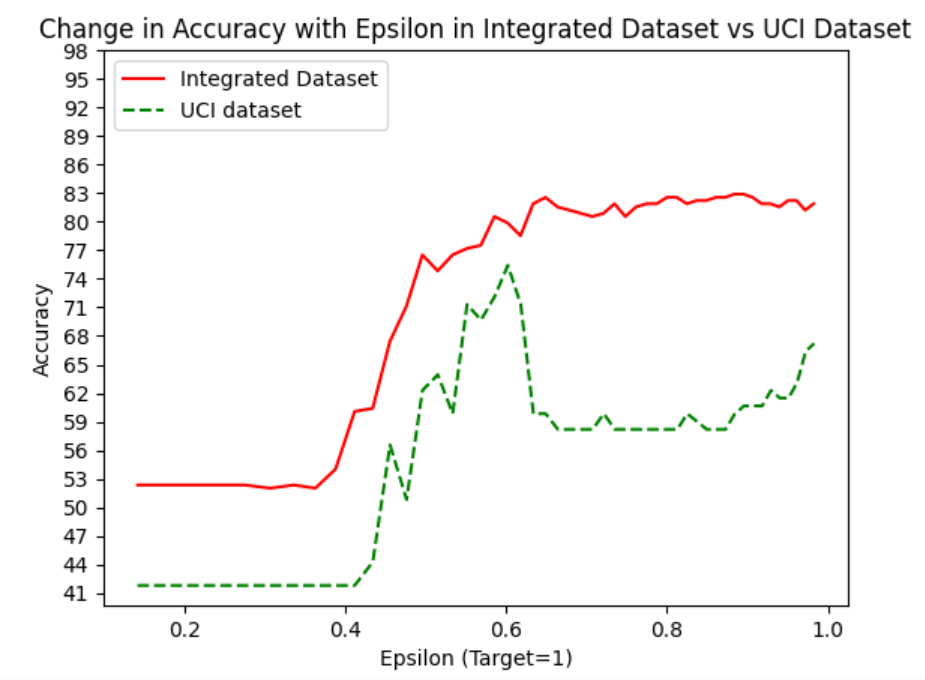}
\caption{Privacy Accuracy Curve Comparison in UCI and Integrated Dataset}
\label{fig::Figure10}
\end{center}\end{figure}

In Fig. \ref{fig::Figure10}, we compare the UCI Dataset with our integrated dataset. The results show that the accuracy curve for the integrated dataset is significantly better, demonstrating superior performance across the training epochs. Under a target epsilon of 1, the integrated dataset achieves higher accuracy and offers better utility while maintaining robust privacy guarantees compared to the UCI dataset. The improved performance of the integrated dataset suggests that combining both datasets enables the model to learn more effectively from a broader and more diverse set of data, leading to better generalization and an improved privacy-accuracy tradeoff. By carefully tuning the privacy parameters epsilon and delta, we were able to strike an optimal balance between privacy and model accuracy. This demonstrates that our framework successfully addresses the trade-off between strong privacy guarantees and maintaining high predictive performance. This balance makes our approach particularly suitable for healthcare applications, where data privacy and accurate predictions are crucial.

\newpage
\section{Conclusion}
\label{sec:con}

This research explores the integration of differential privacy and federated learning for privacy-preserving healthcare analytics, particularly in heart disease prediction. We implemented differential privacy using the Opacus library and evaluated its impact on deep learning models trained on heart disease datasets. Our findings highlight the trade-off between privacy and model accuracy, emphasizing the importance of optimizing privacy parameters to maintain data utility while ensuring security. Experiments with dataset integration and different training configurations demonstrated that combining datasets enhances model performance and stability, improving the privacy-accuracy tradeoff. Additionally, differential privacy was found to mitigate overfitting, further strengthening model generalization across different datasets. Our approach ensures decentralized, privacy-preserving data analysis while maintaining robust predictive performance, making it highly suitable for sensitive healthcare applications. Future work will focus on refining differential privacy techniques and exploring advanced federated learning frameworks to further enhance privacy and accuracy in medical AI systems.

\end{document}